\documentclass[10pt,twocolumn,letterpaper]{article}

\usepackage[pagenumbers]{wacv} 

\usepackage[ruled,vlined]{algorithm2e}
\usepackage{amsmath}  
\usepackage{multirow} 
\usepackage{makecell}
\usepackage{mdframed}
\usepackage{fvextra}   
\newmdenv[backgroundcolor=gray!6,linecolor=gray!55,linewidth=0.6pt,
          roundcorner=4pt,
          innerleftmargin=7pt,innerrightmargin=7pt,
          innertopmargin=4pt,innerbottommargin=4pt,
          frametitlebackgroundcolor=gray!22,
          frametitlefont=\bfseries\small,
          frametitlerule=false,
          splittopskip=10pt]{promptbox}
\usepackage{xcolor}
\usepackage[table]{xcolor}

\usepackage[accsupp]{axessibility}  

\newcommand{\vx}{\mathbf{x}}
\newcommand{\mG}{\mathbf{G}}

\newcommand{\mV}{\mathbf{V}}
\newcommand{\sS}{\mathbb{S}}

\newcommand{\ourslong}{\textbf{Think Before Watching}\xspace}
\newcommand{\oursshort}{\textbf{TBW}\xspace}
\newcommand{\myparagraph}[1]{\vspace{0pt}\noindent{\bf #1}}

\definecolor{wacvblue}{rgb}{0.21,0.49,0.74}
\usepackage[pagebackref,breaklinks,colorlinks,allcolors=wacvblue]{hyperref}

\def\wacvPaperID{22} 
\def\confName{WACV}
\def\confYear{2027}

\title{Think Before Watching: Planning to Select Frames for Video Understanding}

\author{Heng Yu, Yue Zhao, Mohammad Abbasi, Juze Zhang, Ehsan Adeli \\
Stanford University\\
{\tt\small \{yuheng,yzz,mabbasi,juze,eadeli\}@stanford.edu}
}

\begin{document}
\maketitle
\begin{abstract}
Video understanding requires models to capture complex spatial and temporal dependencies across long sequences; yet, processing every frame is computationally expensive and often redundant.
We present Think Before Watching (\oursshort), a training-free, planning-guided frame-selection framework that enhances video understanding without modifying the underlying model.
Before feeding frames into a vision-language model, an LLM planner sequentially decides which temporal region to inspect next and when to stop.
A VLM executor evaluates each proposed region for question-relevant evidence.
By thinking before watching the video, our method avoids wasting computation on uninformative segments and focuses processing on useful evidence.
\oursshort is plug-and-play in existing video models without any additional training.
Extensive experiments demonstrate that our approach improves accuracy across diverse video-understanding tasks, including long- and short-term reasoning, egocentric and exocentric settings, and both spatial and temporal question types, with the largest gains where the evidence is sparse and must be located.
Most notably, Qwen3-VL-32B equipped with \oursshort achieves an absolute gain of 11.5 points (27.8\% relative) on LVBench.
At a 0.5-point accuracy drop, the full \oursshort pipeline reduces effectively processed frames by $23\times$ and doubles throughput relative to dense fixed-FPS sampling.
Grid-based frame packing lowers the cost of each observation, while planning reduces the number of observations required.

\end{abstract}
    
\section{Introduction}
\label{sec:intro}

Understanding long-form videos requires complex reasoning about the temporal dynamics, spatial relations, and cross-modality cues that unfold over time.
Unlike short clips, crucial informative segments are scattered in the long-form video and interspersed with redundant, uninformative parts.
As a result, sparse sampling inevitably misses some crucial clues.
Meanwhile, processing all frames uniformly provides dense coverage, but it is computationally inefficient and prohibitive.
Selecting a compact yet informative subset of frames appears a paradox: selecting too few frames risks losing critical semantic cues, while processing all frames defeats the very purpose of selection.
An ideal model must somehow decide what is essential without seeing everything first.


\begin{figure}[tb]
    \centering
    \includegraphics[width=0.4\textwidth]{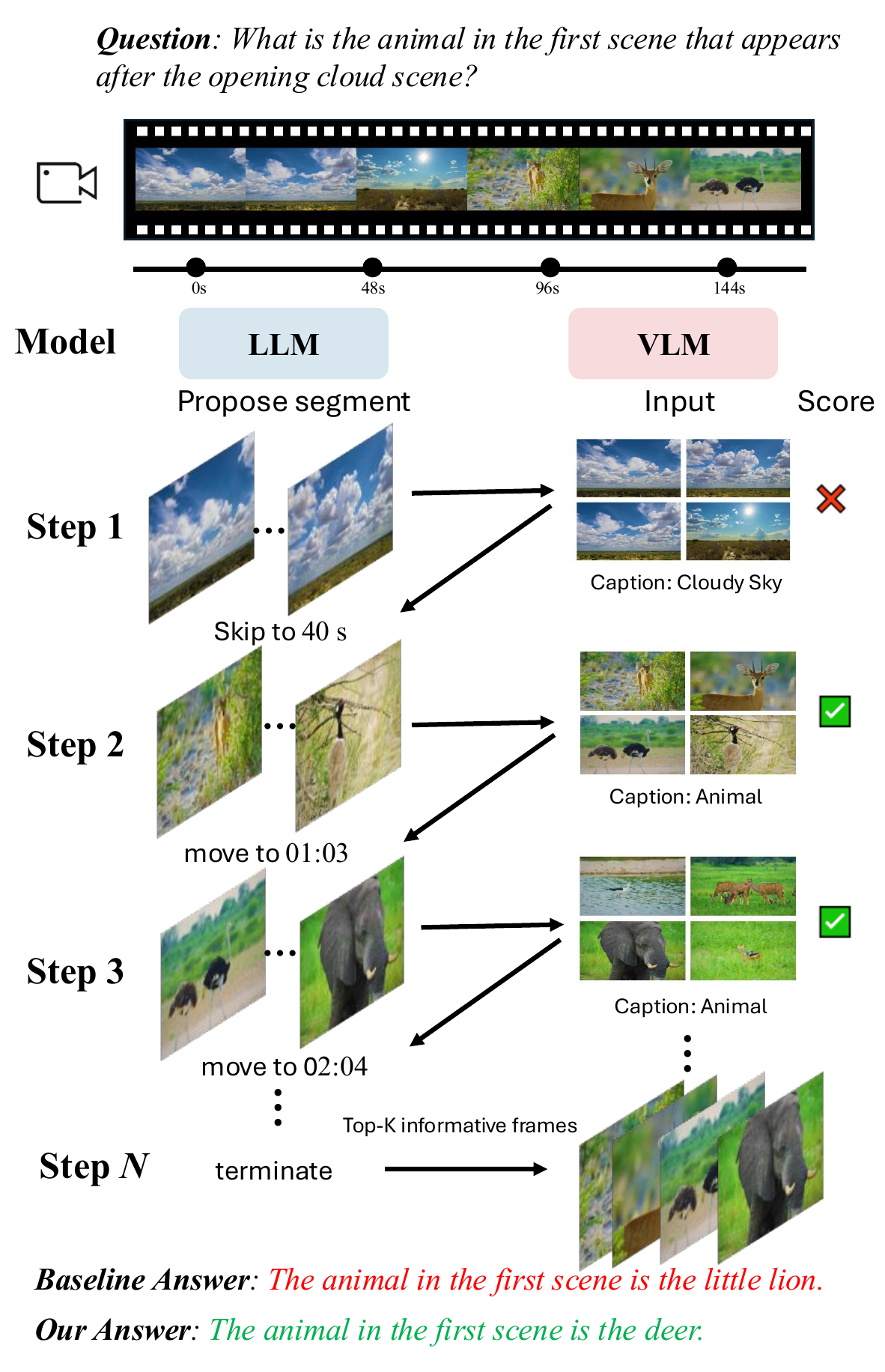}
    \caption{\textbf{Think Before Watching (TBW).} A training-free, planning-guided frame-selection pipeline: a planner devises a search strategy and an executor retrieves informative frames for the VLM, focusing compute on evidence-bearing moments and skipping uninformative segments.}
    \label{fig:teaser}
\end{figure}

In this paper, we formulate frame selection as planning-guided sequential evidence acquisition and leverage the long-context reasoning capability of Large Language Models (LLMs) to \textit{think before watching}.
Our framework includes an LLM planner and a VLM executor:
The LLM planner takes as input the query question and what it already knows about the video, and outputs the timestamp to watch in the next step.
The VLM executor skims through a group of frames, which begins from the timestamp proposed by the LLM planner, and outputs the textual description and the relevance score \wrt the question, which augments the observed states of the LLM planner.
The interaction between the planner and executor continues until the LLM planner collects sufficient cues from observed frames to answer the question and emits a termination signal.
Finally, we select frames from the observed history and make the final prediction.

\oursshort differs from previous approaches in three aspects.
(1) It performs question-conditioned sequential evidence acquisition rather than scoring candidate frames independently~\cite{liang2024end, ye2025fit}.
(2) Its planner adaptively chooses where to inspect next and when to terminate conditioned on the accumulated visual evidence~\cite{wang2024videoagent, yang2025vca}.
(3) It requires no task-specific training and applies to different downstream VLMs without parameter updates.

We conduct extensive experiments on a wide range of benchmarks across diverse video-understanding tasks, including long- and short-term reasoning, egocentric and exocentric settings, and both spatial and temporal question types.
Our method achieves favorable accuracy--efficiency trade-offs compared to existing frame-selection approaches. TBW substantially boosts Qwen3-VL-32B, improving LVBench accuracy from 41.3 to 52.8 (+27.8$\%$ relative), the highest LVBench accuracy among the open-source systems compared in \Cref{tab:main_results}. In addition, TBW delivers consistent gains across Video-MME, VSI-Bench, LongVideoBench, and EgoSchema, demonstrating its robustness and generality.
Qualitative analyses show that the LLM planner exhibits coherent reasoning patterns.
We believe this proposed framework opens new avenues for long-form video understanding through adaptive, planning-guided evidence acquisition.

\section{Related Work}
\label{sec:relatedwork}

\myparagraph{Learned Policies for Frame Selection.}
A major line of work learns a dedicated selector via supervision or reinforcement learning. Early RL-based methods addressed captioning and summarization~\cite{chen2018picknet,liu2022video,rochan2019video}. Recent approaches optimize policy networks for flexible subsampling~\cite{buch2025ffs} or learn to rank candidate frame subsets conditioned on the query~\cite{yu2025framevoyager}, and train a lightweight selector with reinforcement learning~\cite{qin2026efficient}. Others integrate feature-importance/redundancy losses AKS~\cite{tang2025adaptive} or leverage MLLM signals~\cite{hu2025m}. While often strong in accuracy, end-to-end training makes transfer across models/tasks costly and limits plug-and-play use.

\myparagraph{Training-Free but Visually Dependent Samplers.}
An alternative avoids training yet still requires at least one \emph{visual forward pass} to compute selection signals over the whole video. Examples include DPP/importance-based selection (MDP3~\cite{sun2025mdp3}, BOLT~\cite{liu2025bolt}), frames-to-clips sampling~\cite{sun2025frames} and \emph{training-free visual token pruning} to sparsify redundant vision tokens~\cite{jiang2025kind,ye2025fit}. Recent training-free selectors further exploit embedding diversity, query-dependent routing, semantic boundaries, or panel packing~\cite{li2026maxinfo, li2026divide, chen2026wavelet, doorenbos2026video, ben2026himu}; we compare against them in the supplement.
Most remain ``watch-first'', computing selection signals over the whole video before discarding candidates; panel packing instead lowers per-observation cost without choosing where to inspect.

\myparagraph{Agentic and Multi-Step Planners, Localization, and Video-RAG}:
Agent-based pipelines (e.g., VideoAgent~\cite{wang2024videoagent}, VCA~\cite{yang2025vca}, AoTD~\cite{shi2025enhancing}) iteratively plan and retrieve with external memory. Retrieval-centric long-video systems perform coarse-to-fine selection via tree/structure~\cite{wang2025videotree} or document-style retrieval~\cite{ma2025drvideo}; VideoRAG over corpora further improves evidence access~\cite{jeong2025videorag}. Temporal localization provides a closely related formulation via natural-language moment retrieval, including DiDeMo~\cite{anne2017localizing}, TALL~\cite{gao2017tall}, 2D-TAN~\cite{zhang2020learning2dtan}, and LGI~\cite{mun2020local}, as well as the TVR dataset for video–subtitle moment retrieval~\cite{lei2020tvr}. 
Complementary lines on text–video retrieval focus on cross-modal matching at scale, e.g., Frozen-in-Time~\cite{bain2021frozen} and CLIP4Clip~\cite{luo2022clip4clip}. 
While powerful, multi-step agents and retrieval-first pipelines typically incur higher inference latency due to sequential calls and memory operations, motivating our plan-before-perceive approach. In contrast to agent-style pipelines such as VideoAgent that rely on an LLM to perform the final video reasoning, our method keeps the VLM as the primary perceiver and uses the LLM only for lightweight temporal planning, avoiding expensive multi-step LLM inference while preserving strong visual grounding, as the VLM has direct access to visual evidence and can reliably resolve fine-grained spatial relationships that caption-based LLM reasoning may miss.

\myparagraph{Foundations and Long-Context MLLM Efficiency}:
Frame selection complements advances in spatiotemporal modeling~\cite{carreira2017quo,wang2018non,feichtenhofer2019slowfast,arnab2021vivit} and video–language pretraining~\cite{sun2019videobert,luo2020univl,wang2022internvideo,miech2019howto100m,xu2016msr,qasim2025dense,li2016tgif,ma2022x,tong2022videomae,wang2024internvideo2}. Long-context scaling via sparse attention/memory and time-aware design e.g., Longformer~\cite{beltagy2020longformer}, 
Big bird~\cite{zaheer2020big}, 
TimeChat~\cite{ren2024timechat}, MovieChat~\cite{song2024moviechat}, Video-XL~\cite{shu2025video}, and the streaming inference pipeline VideoLLM-Online~\cite{chen2024videollm} is largely orthogonal and complementary: \textbf{TBW} can act as a plug-and-play \emph{pre-processor} that reduces perceptual load regardless of the downstream backbone.

\section{Notations and Preliminaries}
\label{sec:notations}

Given a video $\mV$ and a question $Q$, our goal is to select a compact set of frames $\sS$, whose size is determined by the capacity or computational budget of the downstream VLM, that contains sufficient visual evidence to answer $Q$.
We denote the input video as $\mV = \{\vx_m \in \mathbb{R}^{H\times W\times C},  m = 1,\cdots,T\}$, where $T$ is the total number of frames.
Assuming that the video duration is $t$ seconds, the original frame rate is $r_o = T/t$.

\begin{figure}[tb]
    \centering
    \includegraphics[width=0.98\linewidth]{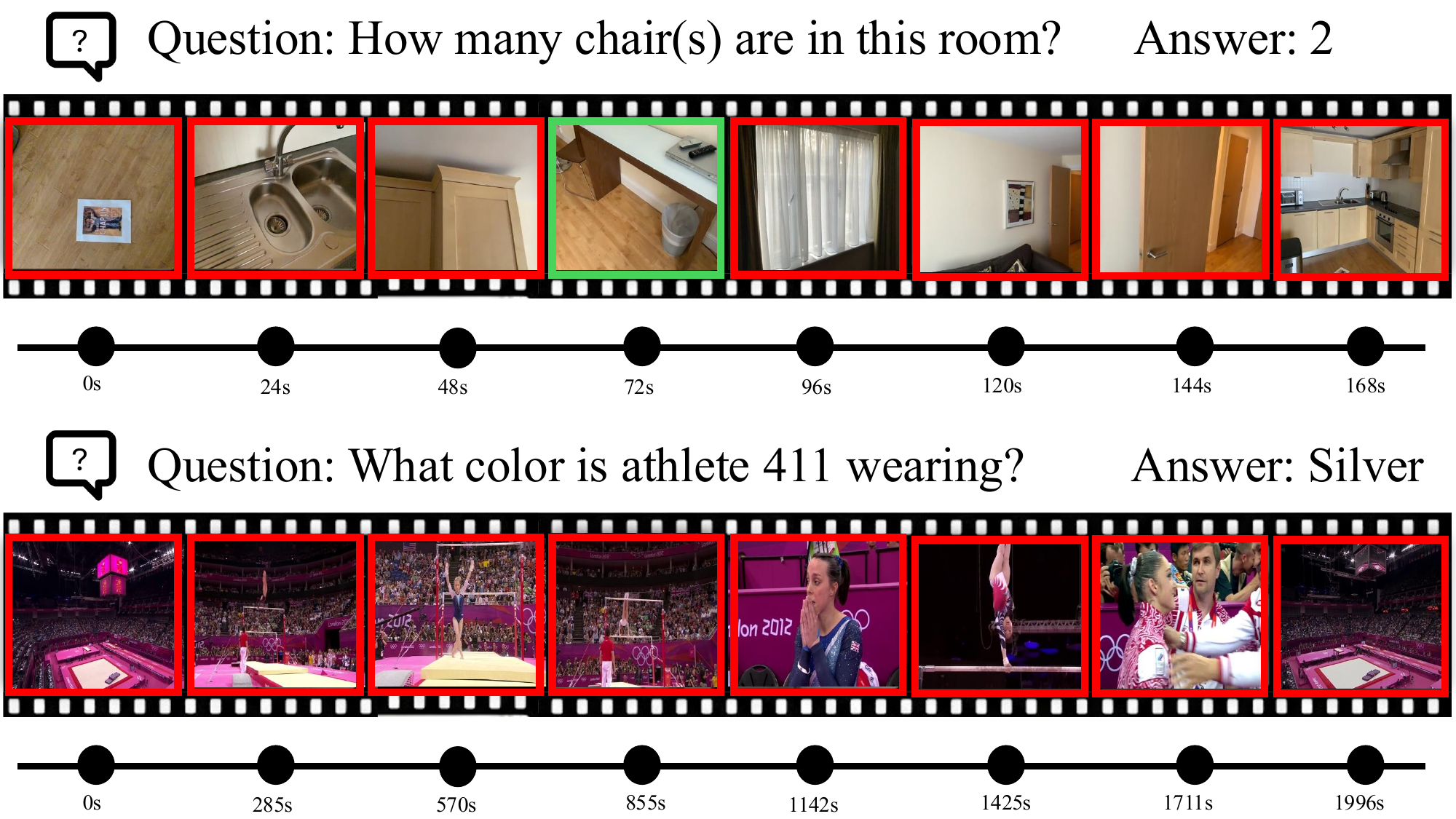}
    \caption{Although sparse uniform sampling is simple and computationally efficient, it often fails to retrieve informative frames that are relevant to the question, since the selected frames may miss key events or visual evidence needed for accurate reasoning.}
    \label{fig:failure}
\end{figure}

\myparagraph{Fixed-number uniform sampling.}
A standard way is to fix the target number of selected frames to be $K$, perform \textit{fixed-number uniform sample} throughout time, and then feed them into the VLM:
\begin{align}
    \sS_\mathit{uniform} = \left\{ \left\lfloor \frac{T}{K}(k+\frac{1}{2}) \right\rfloor, k=0,\cdots,K-1  \right\}.
\end{align}
Although simple, this strategy has a few flaws.
First, uniform sampling is question-agnostic, which often results in irrelevant frames and wasted computation. 
Intuitively, if a question specifically asks about a certain time period, there is no reason to look at the intervals beyond the time period.
Second, uniform sampling is visually agnostic, leading to visually inferior frames due to motion blur or poor composition.
In the worst case, the retrieved frame may be entirely uninformative (\Cref{fig:failure}).

\myparagraph{Fixed-FPS uniform sampling.} Given enough computation and context length, another way is to perform \textit{uniform FPS sampling} with a fixed $\mathit{FPS}=r$,~\eg $r=1$.
\begin{align}
    \sS_\mathit{fps} = \left\{ \left\lfloor r_0(k+\frac{1}{2}) \right\rfloor, k=0,\cdots,\lfloor \frac{r}{r_o}\cdot T \rfloor -1  \right\}.
\end{align}
Despite coverage guarantees of the entire video, it also suffers from the issues mentioned above, leading to linearly increasing computation.



\myparagraph{Selective sampling.}
Beyond uniform sampling, another line of work leverages lightweight detectors such as CLIP~\cite{radford2021learning} or SigLIP~\cite{zhai2023sigmoid}, or even a VLM itself~\cite{wang2024videoagent, arnab2025temporal}, to score frames and retain those deemed relevant to the question.
Although more selective, these approaches typically process frames independently, ignoring temporal dependencies and cross-frame relationships that may be crucial for reasoning. Moreover, without an explicit planning mechanism, they often require traversing the entire video to compute relevance scores, leading to unnecessary computation when only a small subset of frames is ultimately needed.

\section{Method}

\begin{figure}[tb]
    \centering
    \includegraphics[width=0.98\linewidth]{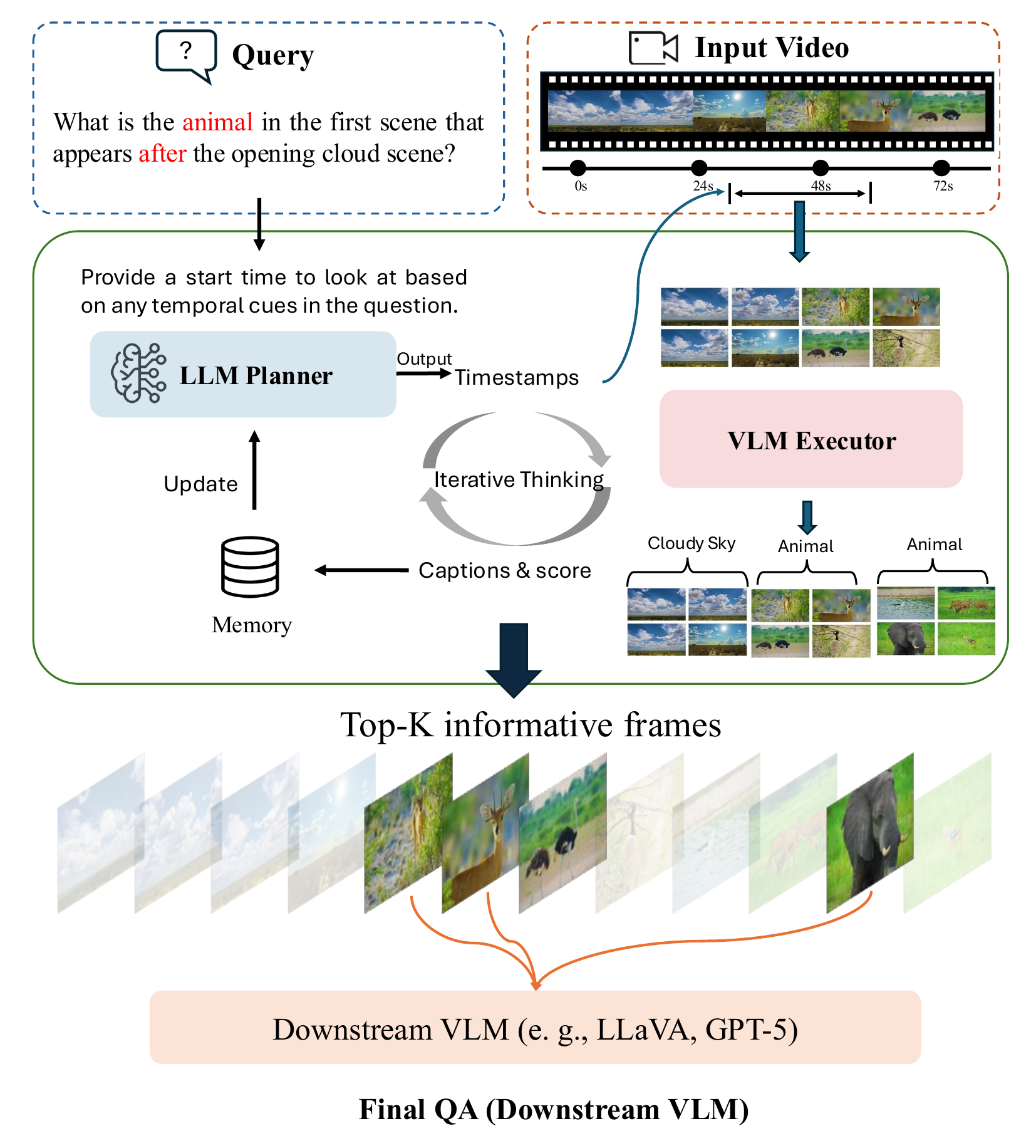}
    \caption{\textbf{Think Before Watching (TBW).} A training-free, planning-guided frame-selection pipeline: a planner devises a search strategy and an executor retrieves informative frames for the VLM, focusing compute on evidence-bearing moments and skipping uninformative segments.}
    \label{fig:method}
\end{figure}

Inspired by recent advances in long-context reasoning in LLMs and VLMs~\cite{beltagy2020longformer, zaheer2020big, ren2024timechat, song2024moviechat, shu2025video}, we propose a planning-guided frame selection framework that jointly employs two complementary modules: 
(1) an \emph{LLM planner} that plans the exploration policy for frame selection,~\eg which frame to start and whether to continue;
and (2) a \emph{VLM executor} that evaluates a group of frames, which is proposed by the LLM planner, for visual evidence regarding the question.
The visual evidence is then passed to the LLM planner.
The LLM planner and the VLM executor run interactively for a few iterations until the planner emits a \textit{termination} signal.
Once done, we gather all frames selected through the reasoning cycle, select a predefined number of them, and pass the final subset to the VLM for answer generation.
The VLM executor can be the same as the final VLM for performance or smaller for efficiency.
\cref{fig:method} provides an overview.


\subsection{LLM as a Planner}
\label{sec:method:planner}

The LLM planner is a reactive, closed-loop controller that decides which temporal segment the video model should inspect next.
Its \textit{state} is the accumulated observation history, its \textit{action} is the next temporal inspection within a bounded range, and it also decides when to terminate.

We formulate the LLM planner as follows.
At each planning step $n$, it takes as input the goal (the query question $Q$) and some state $S_n$ (what the current LLM knows), and outputs an action $A_n$,~\ie
\begin{align}
    A_n = \mathrm{Planner}(Q, S_n).
\end{align}

\myparagraph{Action.}
The action $A_n$ refers to the temporal offset to skip, measured from the end of the previously inspected segment.
Concretely, the action takes value from a discrete set, $A_n\in\{0,\delta,2\delta,\cdots,R_\mathrm{max}\cdot\delta, \langle \texttt{END} \rangle\}$, where $\delta$ is a predefined temporal interval and $R_\mathrm{max}\cdot\delta$ is the maximum range over which the video model can skip.
The action $A_n=0$ corresponds to continuing with the immediately following segment without skipping.
$\langle \texttt{END} \rangle$ means that the LLM planner considers that the goal of gathering sufficient details to answer the question has been reached.
To get the initial time $t_0$, we first try inferring it from the question $t_0 = \mathrm{Planner}(Q)$.
If not, $t_0$ will by default be the beginning of the video, $t_0=0$.

\myparagraph{State.}
The state $S_n$ maintains the full inspection history: the timestamps, textual descriptions, and binary relevance signals of all previously inspected segments,~\ie
\begin{align}
    \label{eq:planner:state}
    S_n = \left\{ (t_j,t_j+\Delta, c_j, h_j) \right\}_{j=0}^{n-1},
\end{align}
where $\Delta$ is the temporal span (in seconds) that the VLM processes at once, $c_j$ denotes the textual descriptions, and $h_j$ is a binary relevance signal.
Given the video snippet being $(t_{n-1},t_{n-1}+\Delta)$, the next timestamp to watch switches to $t_{n} = t_{n-1} + \Delta + A_n$, so that $A_n=0$ yields a contiguous scan and larger actions skip intervening content. 
Both $c_j$ and $h_j$ are outputs of the VLM executor, which will be discussed in the next section.



\subsection{VLM as an Executor}
\label{sec:method:vlm}

\myparagraph{Fast scanning with gridded frames.}
The VLM takes the action signal from the LLM planner and processes the video chunk spanning $[t_j, t_j+\Delta)$, whose $L$ frames start at index $i_j$.
However, this leads to a computational cost similar to the aforementioned uniform FPS sampling, which is too slow in practice.
To enable the executor to run fast, we borrowed the adaptive zooming-in idea~\cite{ye2025re} and reorganized the video frames into a grid:
The sequence of frames is first reshaped into a shape of $N_r \times N_c$, where $L = N_r \cdot N_c$ is the number of frames packed into one grid, and then resized to the regular size.
At sampling rate $r$, a grid therefore spans $\Delta = L/r$ seconds, and the effective number of frames is reduced by $L$ times.
\begin{align}
    [ \vx_{i_j}, \cdots, \vx_{i_j+L-1} ]
    &\rightarrow
    \mG_j = \begin{bmatrix}
    \vx_{i_j}  &  \cdots  &  \vx_{i_j+N_c-1} \\
    \vdots &  & \vdots \\
    \vx_{i_j+L-N_c} &  \cdots  &  \vx_{i_j+L-1} \\
    \end{bmatrix} 
    \notag \\
    &\rightarrow
    \mG_j' = \mathrm{Resize}(\mG_j, (N_r, N_c)).
\end{align}
The VLM takes as input each resized grid $\mG_j'$ and the question $Q$ and outputs a tuple 
\begin{align}
    (h_j, c_j) = \mathrm{VLM}(\mG'_j, Q),
\end{align}
where $c_j$ is a concise caption that describes the grid and $h_j \in \{0,1\}$ is a binary relevance signal, indicating whether the current video segment is relevant to answer the question\footnote{We also studied finer-grained scoring schemes, but observed no significant difference.}.
Each inspection appends $(t_j, t_j+\Delta, c_j, h_j)$ to the planner state in~\cref{eq:planner:state}; segments marked irrelevant ($h_j=0$) are retained, since they inform the planner where not to look again.

\myparagraph{Final prediction with selected frames.}
When the LLM planner reaches the end, we retain only the inspected segments marked relevant ($h_j{=}1$) and subsample from them a predefined number $K$ of frames--determined by the downstream VLM’s capacity or computational budget--balancing question relevance against temporal coverage.
If more than $K$ relevant segments are available, we uniformly sample $K$ of them and represent each by its central frame.
Otherwise, we use all relevant segments and fill the remaining budget by randomly sampling from the temporal intervals not yet covered.

\section{Experiments}

\subsection{Experimental Setup}

\myparagraph{Benchmarks.}
We evaluate our framework on five representative benchmarks that collectively span
long-form, egocentric, and spatial--temporal reasoning in video understanding.

\noindent\textit{(1) LongVideoBench}~\cite{wu2024longvideobench} focuses on long-context video-language understanding,
featuring 3,763 web-collected videos (up to one hour) with 6,678 human-annotated multiple-choice questions.
It introduces a ``referring reasoning'' paradigm that requires retrieving and reasoning over specific visual contexts, addressing the single-frame bias of prior short-video datasets.

\noindent\textit{(2) LVBench}~\cite{wang2024lvbench} is a large-scale benchmark for long-form video reasoning covering
diverse domains such as movies, sports, and documentaries, evaluating complex temporal and causal reasoning across videos lasting tens of minutes to hours.

\noindent\textit{(3) VSI-Bench}~\cite{yang2025thinking} targets spatial-temporal reasoning in egocentric settings,
containing over 5{,}000 QA pairs grounded in 288 indoor videos from 3D reconstruction datasets,~\eg ScanNet and ARKitScenes, and focuses on geometric and object-centric reasoning within embodied environments.

\noindent\textit{(4) VideoMME}~\cite{fu2024video} is a comprehensive multimodal benchmark containing
900 videos spanning 254 hours and 2,700 annotated Q\&As across 6 major domains and 30 sub-domains.
It evaluates robustness across temporal scales, from 11-second clips to hour-long videos, requiring comprehension of both local visual detail and long-range context.

\noindent\textit{(5) EgoSchema}~\cite{mangalam2023egoschema} comprises 5,031 egocentric videos whose multiple-choice questions require temporal ordering, causal inference, and intention prediction from first-person views, testing long-term temporal coherence over realistic daily activities.

\myparagraph{Metrics.}
We report mean relative accuracy for VSIBench~\cite{yang2025thinking} and standard accuracy for the remaining four benchmarks.
In addition, to compare \textbf{efficiency}, we propose the concept of \textbf{effective frame cost (EFC)}, which is the number of frames processed by fast scanning in \Cref{sec:method:vlm}.
\begin{align}
    \text{EFC} = |S_n| = \frac{N}{N_r\cdot N_c},
    \label{eq:efc}
\end{align}
where $N$ denotes the total number of ``seen'' frames.

\myparagraph{Implementation Details.}
We employ both large language models (LLMs) and vision–language models (VLMs) in our framework.
For the LLM, we use Qwen3~\cite{yang2025qwen3}.
For the VLMs, we evaluate two representative open-source models: Qwen3-VL~\cite{team2025qwen3} and InternVL3.5~\cite{wang2025internvl3} .
Together, these models represent the current state of the art among open-source VLMs, providing fair and transparent baselines for comparison with proprietary systems.
All experiments were conducted on four NVIDIA L40S GPUs.


\subsection{Main results}

\begin{table*}[!tb]
\centering
\caption{
\small{Video question-answering accuracy (\%) of various MLLMs across different video understanding benchmarks. Entries with the suffix ``(w/. \oursshort)'' use frames selected by our planning-guided frame selection method, while others adopt the default sampling strategy suggested by the system. final frame budgets ($K$) are shown in gray parentheses after each score, e.g., {\scriptsize\textcolor{gray}{(64)}}. 
Numbers starting with $\sim$ mean that the model uses the fix-FPS uniform sampling (\Cref{sec:notations}) and we report the average frames used.
\textbf{Size} denotes the language model size.
\textbf{V-MME} = Video-MME, \textbf{VSI} = VSIBench, \textbf{LVB} = LongVideoBench.
}}
\resizebox{0.8\linewidth}{!}{
\begin{tabular}{lcllllll}
\toprule
\textbf{Model} & \textbf{Size} & \textbf{LVBench} & \textbf{V-MME} & \textbf{VSI} & \textbf{LVB} & \textbf{EgoSchema} \\
\midrule
GPT-4o~\cite{hurst2024gpt}             & N/A  & \underline{48.9} {\scriptsize\textcolor{gray}{(60)}} & 71.9 {\scriptsize\textcolor{gray}{(384)}} & 47.8 {\scriptsize\textcolor{gray}{(64)}} & \underline{66.7} {\scriptsize\textcolor{gray}{(256)}} {\scriptsize\textcolor{gray}{}} & -- \\
Gemini-1.5-Flash~\cite{team2024gemini}    & N/A  & --  & 70.3 {\scriptsize\textcolor{gray}{($\sim$600)}} & 45.7 {\scriptsize\textcolor{gray}{(n/a)}} & 61.6 {\scriptsize\textcolor{gray}{(256)}}   & 65.7 {\scriptsize\textcolor{gray}{($\sim$180)}} \\
Gemini-1.5-Pro~\cite{team2024gemini}      & N/A  & 33.1 {\scriptsize\textcolor{gray}{($\sim$3.6k)}}   & \underline{75.0} {\scriptsize\textcolor{gray}{($\sim$600)}} & 48.8 {\scriptsize\textcolor{gray}{(n/a)}} & 64.0 {\scriptsize\textcolor{gray}{(256)}} {\scriptsize\textcolor{gray}{}} & 72.2 {\scriptsize\textcolor{gray}{($\sim$180)}} \\
InternVL2.5~\cite{chen2024expanding}   & 72B & 43.6 {\scriptsize\textcolor{gray}{(16)}}    & 72.1 {\scriptsize\textcolor{gray}{(64)}}   & --  & --  & --  \\
InternVL2~\cite{chen2024far} ~\cite{chen2024internvl}    & 40B & 39.6 {\scriptsize\textcolor{gray}{(16)}}    & 61.2 {\scriptsize\textcolor{gray}{(16)}}    & 37.0 {\scriptsize\textcolor{gray}{(32)}}  & 59.3 {\scriptsize\textcolor{gray}{(16)}}  & --  \\
LLaVA-Video~\cite{zhang2024llavavideo}      & 72B & --    & 70.6 {\scriptsize\textcolor{gray}{(64)}}  & 40.9 {\scriptsize\textcolor{gray}{(32)}}  & 61.9 {\scriptsize\textcolor{gray}{(128)}}  & 65.6 {\scriptsize\textcolor{gray}{($\sim$180)}}  \\
VideoLLaVA~\cite{lin2024video}           & 7B  & --     & 39.9 {\scriptsize\textcolor{gray}{(8)}}    & --   & 39.1 {\scriptsize\textcolor{gray}{(8)}}   & --   \\
TimeMarker~\cite{chen2024timemarker}         & 8B  & 41.3 {\scriptsize\textcolor{gray}{(128)}}    & 57.3 {\scriptsize\textcolor{gray}{(128)}}   & --  & 56.3 {\scriptsize\textcolor{gray}{(128)}}   & --   \\
\midrule
InternVL3.5~\cite{wang2025internvl3}           & 8B  & 37.5 {\scriptsize\textcolor{gray}{(8)}}    & 58.4 {\scriptsize\textcolor{gray}{(8)}}    & 50.8 {\scriptsize\textcolor{gray}{(8)}}  & 56.8 {\scriptsize\textcolor{gray}{(8)}}  & 60.4 {\scriptsize\textcolor{gray}{(8)}}  \\
InternVL3.5 (w/. \oursshort) & 8B  & 45.1 {\scriptsize\textcolor{gray}{(8)}}    & 62.6 {\scriptsize\textcolor{gray}{(8)}}    & 52.4 {\scriptsize\textcolor{gray}{(8)}}  & 58.3 {\scriptsize\textcolor{gray}{(8)}}  & 62.8 {\scriptsize\textcolor{gray}{(8)}}  \\
Qwen3-VL~\cite{team2025qwen3}             & 8B  & 33.5 {\scriptsize\textcolor{gray}{(8)}}    & 53.9 {\scriptsize\textcolor{gray}{(8)}}   & 50.4 {\scriptsize\textcolor{gray}{(8)}}  & 54.2 {\scriptsize\textcolor{gray}{(8)}}  & 66.2 {\scriptsize\textcolor{gray}{(8)}}  \\
Qwen3-VL (w/. \oursshort)    & 8B  & \underline{48.9} {\scriptsize\textcolor{gray}{(8)}}    & 62.1  {\scriptsize\textcolor{gray}{(8)}}  & \underline{56.4} {\scriptsize\textcolor{gray}{(8)}}  & 60.8 {\scriptsize\textcolor{gray}{(8)}}  & 67.7 {\scriptsize\textcolor{gray}{(8)}}  \\
Qwen3-VL            & 32B  & 41.3 {\scriptsize\textcolor{gray}{(32)}}    & 72.5 {\scriptsize\textcolor{gray}{(64)}}   & 61.1 {\scriptsize\textcolor{gray}{(32)}} & 64.2 {\scriptsize\textcolor{gray}{(32)}}  & \underline{72.3} {\scriptsize\textcolor{gray}{(32)}} \\
Qwen3-VL (w/. \oursshort)    & 32B  & {\bf 52.8} {\scriptsize\textcolor{gray}{(32)}}    & {\bf 76.1} {\scriptsize\textcolor{gray}{(64)}}  & {\bf 62.6} {\scriptsize\textcolor{gray}{(32)}}  & {\bf 68.1} {\scriptsize\textcolor{gray}{(32)}}  & {\bf 73.2} {\scriptsize\textcolor{gray}{(32)}}  \\
\bottomrule
\end{tabular}
}
\label{tab:main_results}
\end{table*}

\myparagraph{System-level comparison.}
We evaluate \ourslong by replacing the default sampling techniques in each VLM system with our proposed planning-guided frame selection.
As shown in Table~\ref{tab:main_results}, \oursshort consistently improves the performance of Qwen3-VL with different model sizes (8B and 32B) across all benchmarks \textit{without any additional training}.
At the same final frame budget, \oursshort narrows the gap to substantially larger proprietary systems and in several settings surpasses them.

\begin{table}[!tb]
    \centering
    \caption{\small{Comparison with representative content-aware frame selection baselines on LVBench. EFC denotes the effective frame cost defined in \cref{eq:efc}.}}
    \label{tab:comp2}
    \setlength{\tabcolsep}{5pt}
    \resizebox{0.9\linewidth}{!}{
    \begin{tabular}{lcccc}
        \toprule
        & CLIP~\cite{radford2021learning} & VideoAgent~\cite{wang2024videoagent} & TCoT~\cite{arnab2025temporal} & Ours \\
        \midrule
        EFC & 4000 & 3000 & 4000 & 172 \\
        Accuracy & 40.1 & 43.7 & 49.4 & 48.9 \\
        \bottomrule
    \end{tabular}
    }
\end{table}

\myparagraph{Comparisons with frame-selection baselines.}
On LVBench, we compare \oursshort with three representative content-aware frame selection baselines that estimate relevance rather than relying on uniform sampling. \textit{(1) CLIP}~\cite{radford2021learning} ranks frames by similarity and selects the top-$K$. \textit{(2) VideoAgent}~\cite{wang2024videoagent} adopts an LLM-based agent that iteratively retrieves additional evidence via CLIP based on missing information. \textit{(3) Temporal Chain-of-Thought (TCoT)}~\cite{arnab2025temporal} uses the VLM itself to identify and aggregate relevant frames, but requires processing all predefined video segments to ensure coverage.
As shown in Table~\ref{tab:comp2}, \oursshort achieves a favorable accuracy–efficiency trade-off, attaining comparable or higher accuracy while inspecting substantially fewer frames.
In particular, \oursshort matches planning-style approaches such as VideoAgent and TCoT with fewer than an order-of-magnitude processed frames.


\myparagraph{Binary gating versus soft ranking.}
We further study whether the executor should output a binary relevance decision or a soft relevance score for frame selection.
While soft scores provide a natural way to rank candidate segments, we find that they are less reliable under grid-based frame packing, where each segment is represented by multiple low-resolution frames and relevance can vary across questions.
In this setting, the executor is mainly used to decide whether a segment contains potentially useful evidence, rather than to produce a calibrated ranking across all segments.
As shown in Table~\ref{tab:score_rebuttal}, using soft scores for either weighted sampling or top-score selection does not improve accuracy over binary gating.
This suggests that a simple binary relevance signal is sufficient for evidence accumulation and early stopping, while leaving fine-grained visual reasoning to the downstream VLM.


\begin{table}[t]
\centering
\small
\setlength{\tabcolsep}{5pt}
\caption{
\small{
Ablation on binary versus soft executor relevance signals.
}}
\label{tab:score_rebuttal}
\begin{tabular}{l l c c}
\toprule
\textbf{Signal} & \textbf{Strategy} & \textbf{LVBench} & \textbf{Video-MME} \\
\midrule
Binary & \makecell[l]{Uniform over\\relevant segments} & 48.9 & 62.1 \\
Soft & Weighted sampling & 48.5 & 61.8 \\
Soft & Top-score selection & 48.8 & 62.0 \\
\bottomrule
\end{tabular}
\end{table}





\subsection{Efficiency analysis}
We explore the efficiency from two aspects: the effective frame cost (EFC) and the runtime latency.

\begin{figure}[tb]
    \centering
    \includegraphics[width=0.4\textwidth]{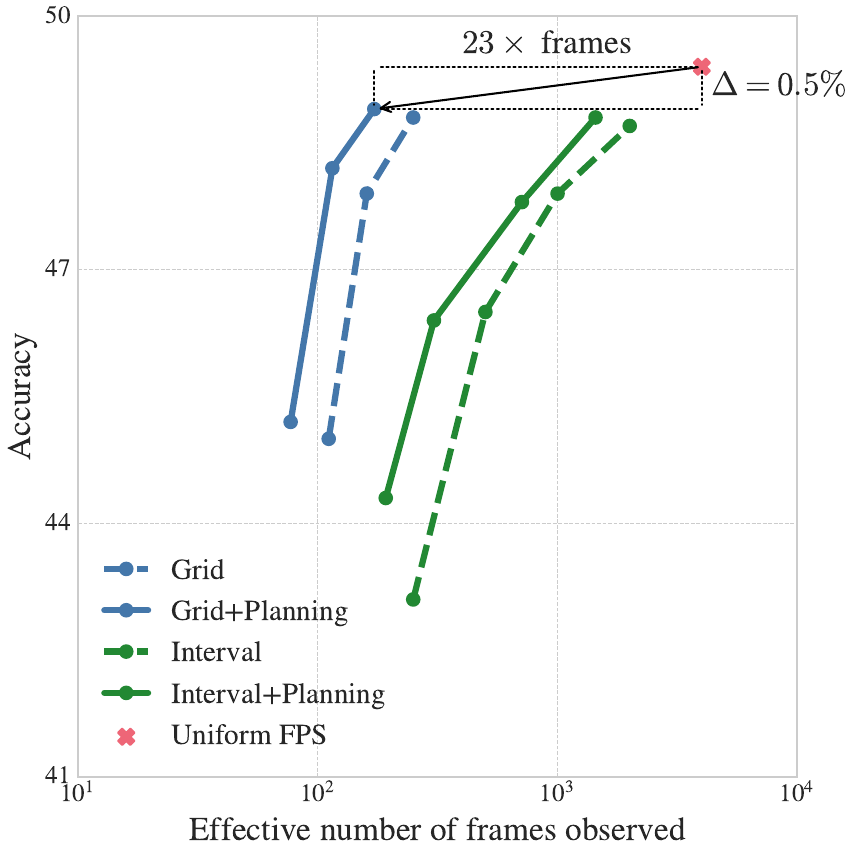}
    \caption{Accuracy-EFC tradeoff on LVBench. 
    \oursshort reduces EFC by $23\times$ ($16\times$ from grid packing and a further $1.45\times$ from planning) while losing only 0.5 accuracy points.}
    \label{fig:compute_accuracy_curve}
    \vspace{-7pt}
\end{figure}

\begin{table}[tb]
    \centering
    \caption{Accuracy-latency trade-off on LVBench. 
    \#Seg denotes the number of temporal segments processed in parallel. 
    Avg. Time (s) reports the average per-video end-to-end inference time.}
    \label{tab:latency}
    \setlength{\tabcolsep}{6pt}
    \resizebox{0.45\textwidth}{!}{
        \begin{tabular}{c c c c c}
            \toprule
            \textbf{Method} & \textbf{\#Seg} & \textbf{EFC} & \textbf{Acc.} & \textbf{Avg. Time} (s) \\
            \midrule
            \multirow{3}{*}{Ours}
                & 1 & 172 & 48.9 & $\sim$35 \\
                & 2 & 189 & 48.7 & $\sim$20 \\
                & 4 & 211 & 48.2 & $\sim$15 \\
            \midrule
            \rowcolor{gray!15}
            Fixed-number
                & -- & 8 & 33.5 & $\sim$5 \\
            \rowcolor{gray!15}
            Fixed-FPS
                & 8 & 4000 & 49.4 & $\sim$70 \\
            \bottomrule
        \end{tabular}
    }
    
\end{table}

\myparagraph{Accuracy-EFC tradeoff.}
\oursshort provides a flexible mechanism to control the effective frame cost (EFC) while maintaining high accuracy.
\cref{fig:compute_accuracy_curve} plots the EFC-accuracy frontier of different strategies based on grid packing and interval subsampling.
The red crossmark corresponds to TCoT~\cite{arnab2025temporal}, which does not skip unseen regions here and therefore scans all predefined segments.
This strategy is inefficient because it must process many irrelevant frames before reaching the evidence, resulting in high effective frame cost.
In contrast, our model shifts the curve dramatically leftward.
On LVBench, we reduce the effective number of frames seen by $23\times$ at a loss of 0.5 accuracy points.
Grid packing accounts for a $16\times$ reduction in the cost of each observation, and planning for a further $1.45\times$ reduction in the number of observations at matched accuracy, with a growing accuracy advantage at tighter budgets (\Cref{tab:grid_ablation}).

The four curves show different configurations.
\textit{(1) With and without planning.}
When LLM planning is disabled, the action space becomes a singleton set,~\ie $A_n=\{0\}$, yielding a contiguous scan over the whole video.
With planning, both tradeoff curves shift further toward the low-cost regime while maintaining competitive accuracy.
\textit{(2) grid-based packing \vs temporal sampling.}
\oursshort benefits from two main technical aspects,~\ie the grid-based packing and a set of temporal intervals to skip (\cref{sec:method:planner}).
To showcase their respective advantages, we add one more setting by using the basic $1\times 1$ grid (the green ``interval''-only curves).
We provide more ablative results in \Cref{sec:results:ablation}.

\myparagraph{Latency.}
The vanilla \oursshort requires a sequential loop between the LLM planner and VLM executor.
This leads to increased latency, which we study in \Cref{tab:latency}.
To alleviate this issue, we devise a simple variant of \oursshort: it first processes multiple temporal segments in parallel, runs independent planner–executor instances, and then merges candidate frames before final VLM inference.
This enables multi-GPU parallelism to reduce wall-clock latency at the cost of higher EFC and a modest accuracy drop.
From \Cref{tab:latency}, we observe that the runtime of \oursshort decreases from 35 seconds per video to 15 when processing 4 segments in parallel.
Meanwhile, the effective frame cost increases by 23\% and accuracy decreases by 0.7 points.
Compared to dense fixed-FPS uniform sampling, \oursshort runs $2\sim4.67\times$ faster with a marginal drop of $0.5$--$1.2$ accuracy points.
Sparse fixed-number sampling remains substantially faster ($\sim$5s), but its LVBench accuracy is 15.4 points lower than sequential \oursshort.

\subsection{Ablation}
\label{sec:results:ablation}

We conduct ablation experiments to understand the effects of three major design factors on LVBench:
(1) grid size for spatial frame packing, 
(2) temporal interval for subsampling, and 
(3) the final number of frames passed to the downstream VLM. 
The first two examine how different compression strategies interact with planning, while the third evaluates whether the frames selected by our planner remain informative as the frame budget increases.

\myparagraph{Effect of grid size.}
\Cref{tab:grid_ablation} studies grid-based packing by increasing the grid size from $1{\times}1$ (no packing) to $6{\times}6$ (highly packed).
Larger grids pack more frames into a single image, which reduces the EFC but simultaneously lowers the resolution available to each sub-frame.
We observe that accuracy remains relatively stable for moderate grid sizes.
For example, increasing the grid from $1{\times}1$ to $4{\times}4$ only incurs minimal degradation (from $49.4\%$ to $48.8\%$).
However, once the grid becomes overly dense (e.g., $5{\times}5$ or $6{\times}6$), accuracy declines more noticeably.
This sharper drop occurs because, under a fixed input resolution, packing too many frames forces each sub-frame to become extremely low-resolution, ultimately impairing the VLM’s ability to interpret fine-grained visual content.
Overall, this monotonic trend shows that grid packing offers a stable and predictable efficiency-accuracy tradeoff.
The two knobs therefore degrade differently: packing lowers the cost of each observation by reducing the resolution of every packed frame, which we attribute the sharp drop beyond $4{\times}4$ to, whereas planning reduces how many observations are made without degrading any of them.
Read at matched cost, \Cref{tab:grid_ablation} shows planning ahead at both budgets: $48.2$ versus $45.0$ near $\mathrm{EFC}\approx115$, and $48.9$ versus $47.9$ near $\mathrm{EFC}\approx170$, so its benefit grows as the budget tightens.
To further isolate the contribution of the planner, we keep the $4{\times}4$ executor, final frame budget, and downstream VLM fixed, and replace the LLM planner with two non-LLM policies, each configured so that its \emph{average} effective frame cost matches ours ($\sim$170). Fixed skipping achieves 45.1\%, random bounded skipping 46.4\%, and the LLM planner 48.9\%. These results suggest that adaptively allocating the budget---where to inspect and when to stop---retains more accuracy than fixed or random traversal at comparable average cost.

\myparagraph{Effect of temporal interval.}
\Cref{tab:interval_ablation} varies the temporal interval from 1 second (dense sampling) to 16 seconds (very sparse).
As expected, larger intervals linearly reduce EFC but also increase the risk of skipping key events.
Without planning, accuracy drops more steeply than for grid packing—for example, from $49.4\%$ at 1s to $43.1\%$ at 16s since subsampling discards temporal information entirely.
When combined with planning, EFC falls substantially at every interval while accuracy is essentially unchanged, and at the sparsest setting planning also recovers accuracy ($44.3$ versus $43.1$ at 16s), shifting the tradeoff curve leftward; interval-based methods nonetheless remain more sensitive than grid-based ones.
This highlights that temporal subsampling is a more aggressive but less stable acceleration mechanism.

\myparagraph{Effect of frame budget.}
To evaluate whether the frames selected by the planner remain informative when the downstream frame budget increases, we test Qwen3-VL under 8, 16, and 32 input frames (\Cref{tab:abla_frms}).
Across all benchmarks, our selected frames yield substantially higher accuracy than uniform sampling at every budget.
The gap remains substantial at every budget, but narrows as the budget grows, from $+8.2$ to $+5.8$ on Video-MME and from $+6.0$ to $+2.6$ on VSI-Bench. This mirrors the trend observed above: planning contributes most when the budget is tight, and matters less once enough frames are available for uniform sampling to cover the relevant evidence on its own.

\begin{table}[t]
\centering
\small
\setlength{\tabcolsep}{5pt}
\caption{
\small{
Video question-answering accuracy (\%) of Qwen3-VL under different frame budgets.
``Ours'' denotes frames selected by our planning-guided frame selection method.
}}
\label{tab:abla_frms}
\resizebox{0.98\linewidth}{!}{
\begin{tabular}{lcccc}
\toprule
\textbf{Model} & \textbf{\#Frames} & \textbf{LVBench} & \textbf{Video-MME} & \textbf{VSIBench} \\
\midrule
\multirow{3}{*}{\makecell{Qwen3-VL\\(uniform)}} 
   & 8  & 33.5 & 53.9 & 50.4 \\
   & 16 & 34.9 & 58.0 & 56.4 \\
   & 32 & 39.7 & 61.7 & 59.2 \\
\midrule
\multirow{3}{*}{\makecell{Qwen3-VL\\(Ours)}} 
   & 8  & 48.9 & 62.1 & 56.4 \\
   & 16 & 51.2 & 64.2 & 59.3 \\
   & 32 & 53.1 & 67.5 & 61.8 \\
\bottomrule
\end{tabular}
}
\vspace{-10pt}
\end{table}


\begin{figure}[!tb]
    \centering
    
    \begin{subfigure}{0.98\linewidth}
        \centering
        \includegraphics[width=\linewidth]{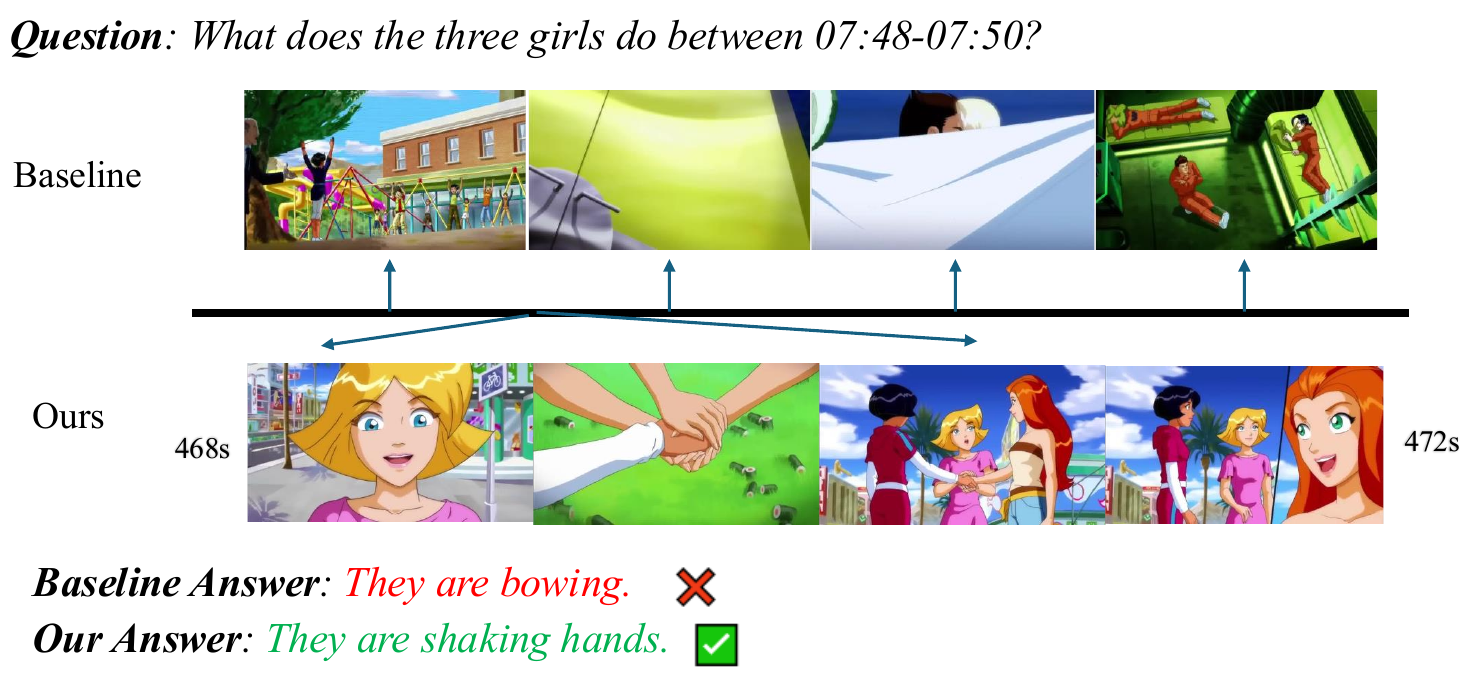}
        \caption{LVBench example}
    \end{subfigure}
    \begin{subfigure}{0.98\linewidth}
        \centering
        \includegraphics[width=\linewidth]{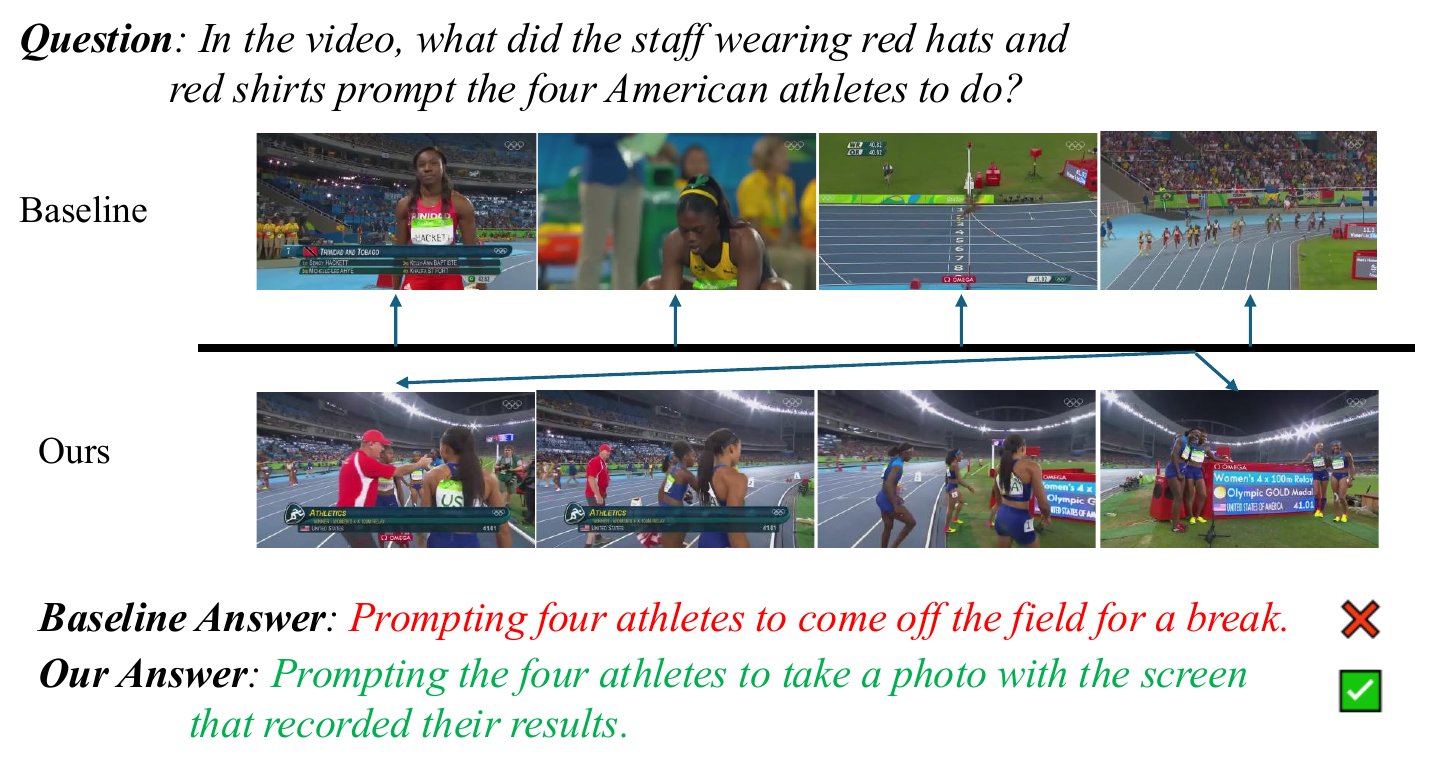}
        \caption{Video-MME example}
    \end{subfigure}
    \begin{subfigure}{0.98\linewidth}
        \centering
        \includegraphics[width=\linewidth]{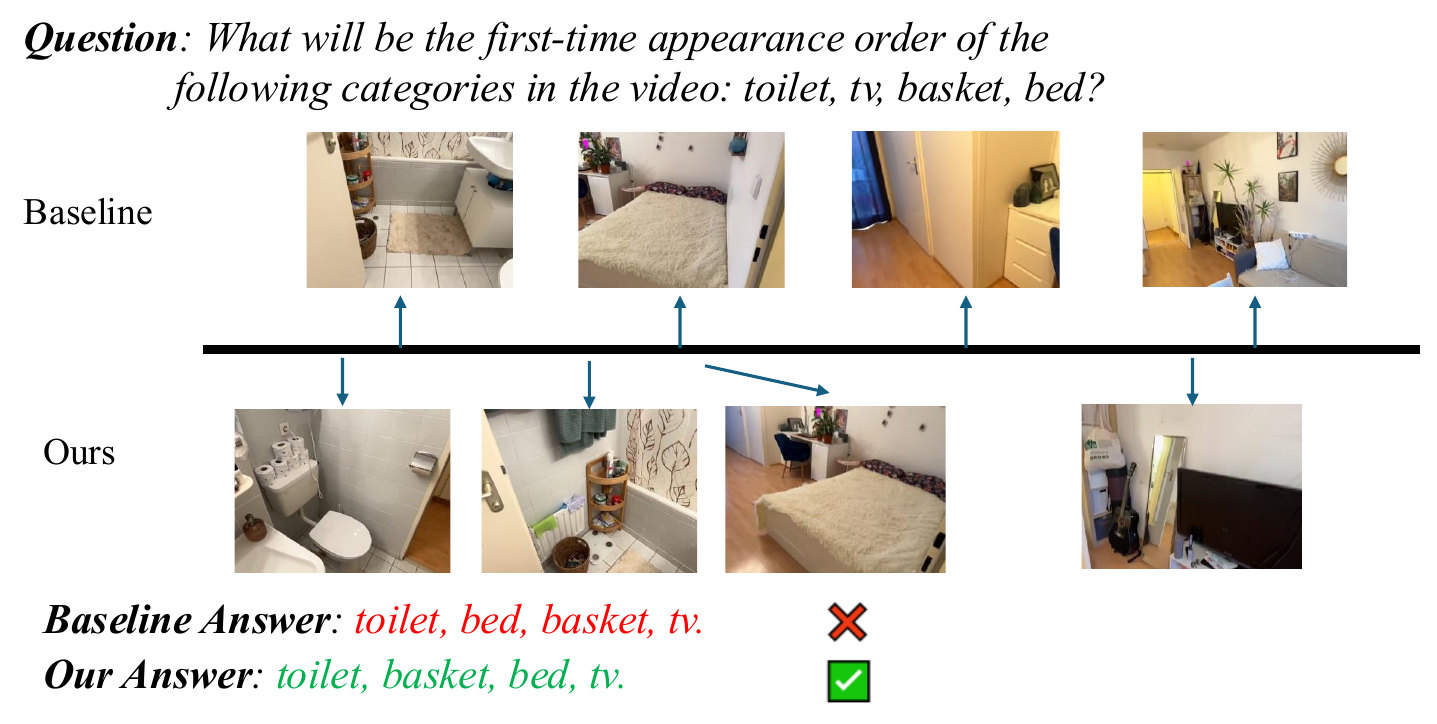}
        \caption{VSI-Bench example}
    \end{subfigure}
    \caption{\small{Comparison of frames sampled by the baseline and the proposed \oursshort, shown respectively for LVBench, Video-MME, and VSI-Bench.}}
    \label{fig:lvbench_qual}
    \vspace{-5pt}
\end{figure}

\subsection{Qualitative results}

\Cref{fig:lvbench_qual} further illustrates how \ourslong selects evidence in a question-aware manner through three representative cases drawn from LVBench, Video-MME, and VSI-Bench:

\myparagraph{Timestamp-aware reasoning}.
These types of questions are common in benchmarks such as LVBench.
When the question contains explicit temporal references, the planner directly navigates to the specified interval, ensuring that the retrieved frames precisely match the queried moment.

\myparagraph{Visual search with early stopping}.
In most benchmarks such as Video-MME, the question does not indicate any timestamp directly.
In the absence of timestamps, the planner instructs the executor to scan the video from the beginning.
However, since the video contains a wide range of shots, most of which are not relevant to the query, the LLM planner can terminate the process early, once the queried action or object is detected,
saving substantial computation while preserving answer fidelity.

\myparagraph{Multi-evidence aggregation via history}.
Some questions in benchmarks like VSI-Bench require collecting multiple pieces of visual evidence that may appear far apart.
Here, the planner maintains an evolving history of discovered entities and continues searching only when necessary, stopping once all referenced items have been observed.

Across all three scenarios, uniform sampling fails to capture the decisive evidence, yielding incorrect answers.
Recent methods, such as Temporal Chain-of-Thought, address this by exhaustively traversing the video, but at the cost of dramatically inflated computation.

\begin{table}[t]
\centering
\small
\setlength{\tabcolsep}{6pt}
\caption{
Ablation on grid-based frame packing. 
Larger grids (from $1{\times}1$ to $6{\times}6$) pack more frames per image, 
leading to lower effective frame cost (EFC) and a gradual drop in accuracy.
}
\vspace{-0.1in}
\begin{tabular}{c|cc|cc}
\toprule
\multirow{2}{*}{\textbf{Grid Size}} & 
\multicolumn{2}{c|}{\textbf{Grid}} & 
\multicolumn{2}{c}{\textbf{Grid + Planning}} \\
& EFC & Acc. & EFC & Acc. \\
\midrule
$1{\times} 1$ & 4000 & 49.4 & 2960 & 49.3 \\
$2{\times} 2$ & 1000 & 49.2 & 750  & 49.0 \\
$3{\times} 3$ & 444  & 48.9 & 351  & 48.8 \\
$4{\times} 4$ & 250  & 48.8 & 172  & 48.9 \\
$5{\times} 5$ & 160  & 47.9 & 115  & 48.2 \\
$6{\times} 6$ & 111  & 45.0 & 77   & 45.2 \\
\bottomrule
\end{tabular}
\label{tab:grid_ablation}
\end{table}

\begin{table}[t]
\centering
\small
\setlength{\tabcolsep}{6pt}
\caption{
Ablation on temporal subsampling intervals. 
Larger intervals (from 1s to 16s) reduce effective frame cost (EFC) but risk skipping 
important events, yielding a sharper accuracy drop than grid-based packing.
}
\vspace{-5pt}
\begin{tabular}{c|cc|cc}
\toprule
\multirow{2}{*}{\textbf{Interval}} &
\multicolumn{2}{c|}{\textbf{Interval}} &
\multicolumn{2}{c}{\textbf{Interval+Planning}} \\
& EFC & Acc. & EFC & Acc. \\
\midrule
1s  & 4000 & 49.4 & 2880 & 49.3 \\
2s  & 2000 & 48.7 & 1440 & 48.8 \\
4s  & 1000 & 47.9 & 710  & 47.8 \\
8s  & 500  & 46.5 & 305  & 46.4 \\
16s & 250  & 43.1 & 192  & 44.3 \\
\bottomrule
\end{tabular}
\label{tab:interval_ablation}
\vspace{-5pt}
\end{table}

\subsection{Failure modes}

The main failure modes of \oursshort are inaccurate segment relevance estimates and premature termination, either of which can cause informative evidence to be missed.
Bounded forward actions prevent oscillatory exploration, and errors affect only which frames are selected rather than pipeline stability; additional analysis is provided in the supplement.
\section{Conclusion}

We presented Think Before Watching (\oursshort), a training-free framework for planning-guided sequential evidence acquisition in long videos.
By adaptively choosing which temporal regions to inspect and when to stop, \oursshort improves existing VLMs across diverse video-QA benchmarks while reducing unnecessary visual processing, with gains that vary by benchmark and question type.
Our results further show that grid packing and planning address complementary bottlenecks: packing reduces the cost per observation, whereas planning reduces the number of observations required.

\section{Acknowledgments}
This project was partially supported by NIH grant R01AG089169 and UST. We thank Mohamed Elhoseiny for his insightful feedback and valuable discussions.
{
    \small
    \bibliographystyle{ieeenat_fullname}
    \bibliography{main}
}


\end{document}